\documentclass[11pt]{article}

\usepackage{microtype} 
\usepackage{booktabs}  
\usepackage{url}  

\usepackage{amsmath}
\usepackage{amsthm}

\usepackage{algorithmicx}
\usepackage{algorithm,algpseudocode}
\usepackage{graphicx}
\usepackage{wrapfig}
\newcommand{\Loss}{\mathcal{L}}
\newcommand{\mlr}{$\hat{\eta}$}

\usepackage{multirow}
\usepackage{multicol}
\usepackage{caption}
\usepackage{tabularx}
\usepackage{subcaption}

\newcommand{\plus}{\scalebox{1}{+}}

\usepackage{mathtools}
\newcommand*{\myovline}[2]{\overbracket[#2][-1pt]{#1}}
\newcommand{\met}{$\myovline{\textit{met}}{0.2pt}$}

\usepackage[finalworkshop]{automl}

\usepackage{natbib}

\title{Simple and Effective Gradient-Based Tuning of Sequence-to-Sequence Models}

\author[1]{\nameemail{Jared Lichtarge}{lichtarge@google.com}}
\author[1]{\nameemail{Chris Alberti}{chrisalberti@google.com}}
\author[1]{\nameemail{Shankar Kumar}{shankarkumar@google.com}}

\affil[1]{Google Research}

\hypersetup{%
  pdfauthor={Jared Lichtarge}, 
  pdftitle={Simple and Effective Gradient-Based Tuning of Sequence-to-Sequence Models},
  pdfsubject={},
  pdfkeywords={}
}

\begin{document}

\maketitle

\begin{abstract}
Recent trends towards training ever-larger language models have substantially improved machine learning performance across linguistic tasks. However, the huge cost of training larger models can make tuning them prohibitively expensive, motivating the study of more efficient methods. Gradient-based hyper-parameter optimization offers the capacity to tune hyper-parameters during training, yet has not previously been studied in a sequence-to-sequence setting. We apply a simple and general gradient-based hyperparameter optimization method to sequence-to-sequence tasks for the first time, demonstrating both efficiency and performance gains over strong baselines for both Neural Machine Translation and Natural Language Understanding (NLU) tasks (via T5 pretraining). For translation, we show the method generalizes across language pairs, is more efficient than Bayesian hyper-parameter optimization, and that learned schedules for some hyper-parameters can out-perform even optimal constant-valued tuning. For T5, we show that learning hyper-parameters during pretraining can improve performance across downstream NLU tasks. When learning multiple hyper-parameters concurrently, we show that the global learning rate can follow a schedule over training that improves performance and is not explainable by the `short-horizon bias' of greedy methods \citep{wu2018}. We release the code used to facilitate further research. 

\end{abstract}

\section{Introduction}\label{sec:introduction}

Finding good hyper-parameter values is critical to achieving good performance across machine learning domains; this has inspired much work into hyper-parameter optimization (HPO) 
(see \citet{Feurer2019}). Traditionally popular HPO methods require running many trials of hyperparameter sets in parallel or sequential training runs \citep{Bengio2012,Snoek12,li2016}. These methods become infeasible as the cost of individual runs increases. This difficulty is exacerbated by recent trends towards larger models \citep{bert_2019,gpt3_2020,meena_2020,palm2022}, which have come to dominate progress on linguistic tasks, yet are only sparsely or indirectly tuned.

The growing field of gradient-based HPO methods offers an alternative to conventional HPO by allowing hyper-parameters to be learned based on a loss function, which can greatly improve over the efficiency of comparing constant values tuned across multiple runs \citep{maclaurin15,pedregosa2016,franceschi_bilevel_2018}\footnote{See Appendix \ref{appendix:related_work}
for a full description of related works.}. Many gradient-based methods additionally allow hyper-parameters to dynamically vary in value over a training run as opposed to only taking static values\footnote{We refer to hyper-parameters which vary over a training run as \textit{dynamic}, and those which are constant as \textit{static.}}. However, most prior work on gradient-based HPO methods has not focused on text-processing, with notable exceptions in \cite{hu2019learning} and \cite{lorraine2020}. This domain mismatch makes it unclear how well these methods may work for the large language model setting.

We present the first study of gradient-based hyper-parameter learning on sequence-to-sequence tasks \citep{sutskever2014seq2seq}. We extend a greedy gradient-based approach that has been applied previously to image classification tasks \citep{luketina2016scalable,wu2018,baydin2017online}, as it is simple, generalizable, and easily extensible. This allows us to apply greedy hyper-parameter learning to a) multiple hyper-parameters simultaneously
and b) experiment across models and tasks. We learn hyper-parameters for momentum and learning rate scaling for Transformer \citep{vaswani2017attention} sequence-to-sequence models for neural machine translation (NMT) and T5 model pretraining \citep{raffel2019exploring}. 

For NMT, we show that hyper-parameter schedules can be learned greedily with minimal tuning across language pairs, and that those learned schedules can be more efficient than Bayesian-optimized tuning and more performant than optimal constant-valued tuning. We demonstrate the absence of `short-horizon bias' while learning momentum, and the benefit of treating momentum as a \textit{dynamic} hyper-parameter. For T5, we show that learning a learning rate scalar alongside momentum changes the behavior of that scalar, improving both the convergence speed and performance of T5 pretraining, gains which are reflected in performance on downstream NLU tasks.

\section{Method}\label{sec:method}

We use a method that allows hyper-parameters to be learned greedily
by gradient descent over the course of training.
Per training step, we perform a bi-level optimization to learn both the model parameters via the training loss, and learned hyperparameters via the \textit{guidance loss}. The \textit{guidance set}
is held-out from the training data to provide the loss by which the hyperparameters are learned.

Let $X$ denote a training dataset and $\Omega$ be a general optimizer function for training a model $\theta$ on $X$, with hyperparameters $\lambda$ and loss function $\Loss_X$. Our training method can be summarized as:
\begin{align}
  g_t &= \nabla_{\theta_t} \Loss_X(\theta_t) &
  \theta_{t+1} &= \Omega(\theta_t, g_t, \lambda_{t}) \nonumber \\
  \hat{g}_t &= \nabla_{\lambda_t} \Loss_H(\theta_{t+1}) &
  \lambda_{t+1} &= \hat{\Omega}(\lambda_t, \hat{g}_t, {\hat{\lambda}}) 
  \nonumber,
\end{align}
where at each time step $t$, the updated model parameters $\theta_{t+1}$ are first computed based on the gradient ($g_t$) of the training loss $\Loss_X$. To compute the guidance loss gradients ($\hat{g}_t$) for the hyperparameters, we calculate the loss $\Loss_H$ of the new model $\theta_{t+1}$ on the guidance set. Finally, the updated hyperparameter values $\lambda_{t+1}$ are obtained based on a meta-optimizer $\hat{\Omega}$ with corresponding meta-hyperparameters ${\hat{\lambda}}$. Thus in every training step, we update both the model parameters and the hyperparameters. The process is formalized in 
Algorithm \ref{algo:guided_learning} in Appendix \ref{appendix:algorithm}.

This method is greedy; the horizon of the guidance objective is limited to a single step. 
\cite{wu2018} showed that greedy methods applied to learning the learning rate can have a bias towards tiny learning rates, which prevent them from learning and achieving good performance over longer horizons (\textit{short-horizon bias}). We will explore the practical consequences of this phenomenon by using this method to learn a learning rate scalar $\alpha$ and momentum $\beta_1$.

\section{Experiments}\label{sec:experiments}

For NMT, we use Transformer models with 121M parameters and the LAMB optimizer \citep{lamb}\footnote{For complete experiment setup details, see Appendix \ref{appendix:setup}.
}. We train on NMT datasets from the WMT19 machine translation task \citep{barrault-etal-2019-findings}. For evaluation, we decode using beam search and report BLEU \citep{papineni-etal-2002-bleu} scores. For T5, we use the \textit{small} configuration (60M parameters) and the Adafactor optimizer \citep{shazeer2018adafactor}. We use the C4 dataset \citep{raffel2019exploring}. We report loss on the C4 development set and the same evaluation criteria as the original T5 paper for downstream tasks. For all hyper-parameter learning experiments, we use the Adam optimizer \citep{adam} with default settings as meta-optimizer, tuning only the meta-learning-rate ${\hat{\eta}}$. For the guidance set, we use a single held-out training batch\footnote{In preliminary experiments, we found no benefit to a larger guidance set.}. As the hyper-parameters must vary within constrained ranges, $\alpha$ is kept positive by an exponential activation function, and $\beta_1$ is constrained between 0 and 1 by a sigmoid.


\subsection{Neural Machine Translation}
\begin{figure*}
\centering
\begin{minipage}{.5\textwidth}
  \centering
  \includegraphics[width=1\linewidth]{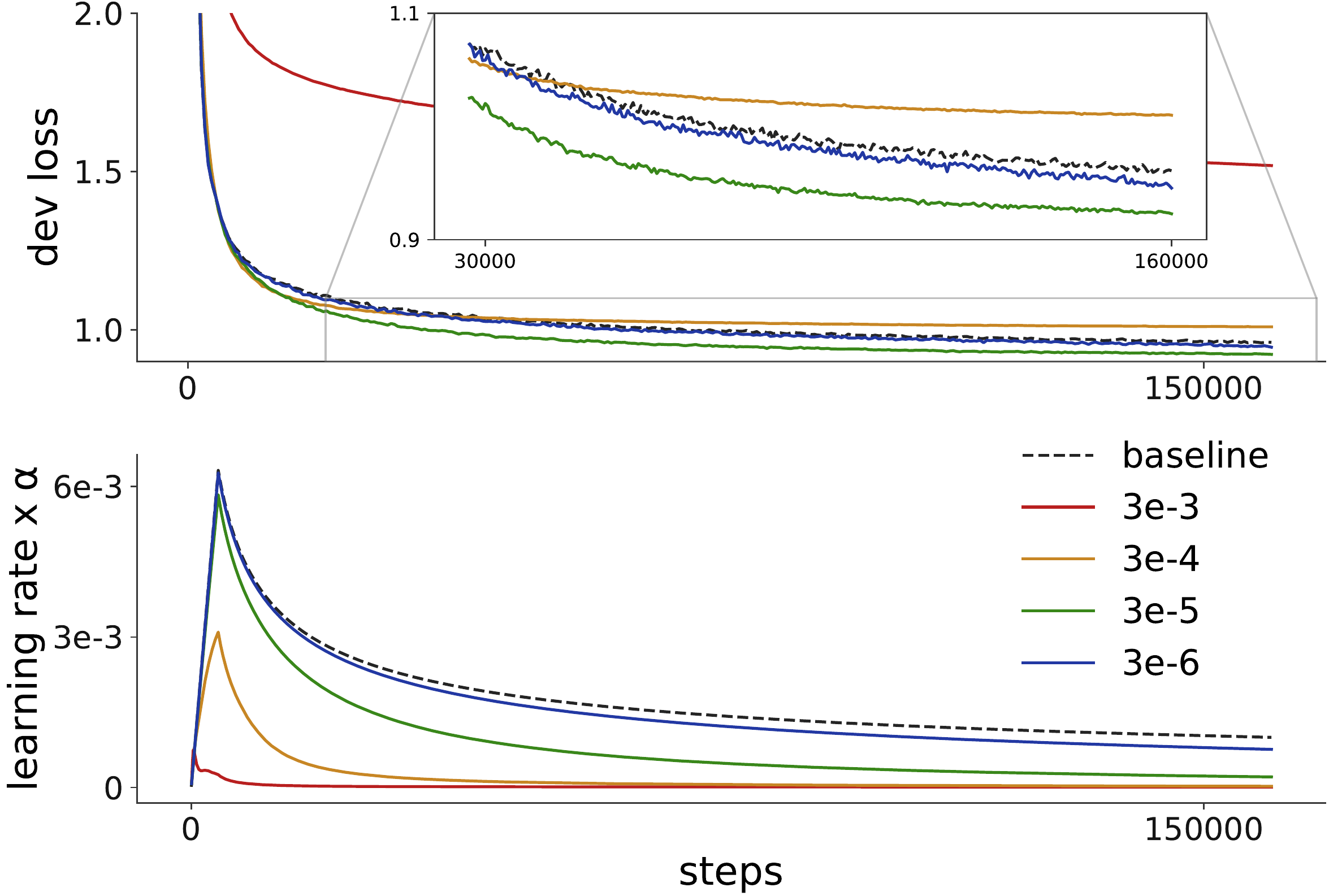}
  \captionof{figure}{Learning $\alpha$, varying \mlr{} values. \\ German-English NMT}
  \label{fig:lr_sweep}
\end{minipage}%
\begin{minipage}{.5\textwidth}
  \centering
  \includegraphics[width=1\linewidth]{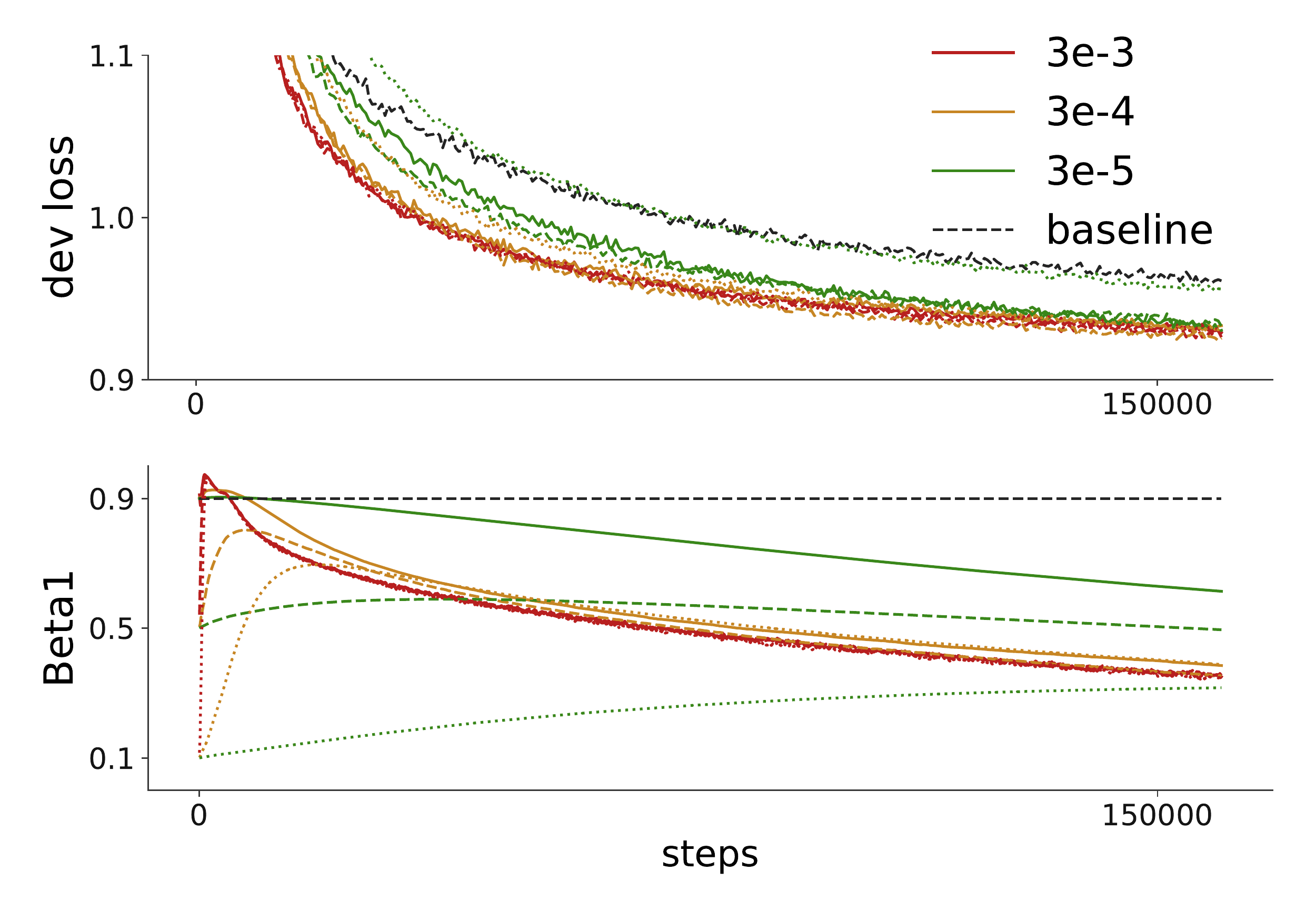}
  \captionof{figure}{Learning $\beta_1$, varying \mlr{} and init. values. \\ German-English NMT}
  \label{fig:beta1_sweep}
\end{minipage}
\end{figure*}

In Figure \ref{fig:lr_sweep}, we compare the evolution of learning the learning rate scalar ($\alpha$) over training for runs with differing meta-learning rates (\mlr) for the German-English language pair. In Figure \ref{fig:beta1_sweep} we do the same for learning $\beta_1$, varying the initialization values in addition to \mlr{}. For learning $\alpha$, the guidance optimization drives all runs to as low a learning rate as is allowed by the meta-learning rate, demonstrating the `short horizon bias'. Note that some guided $\alpha$ runs do outperform the baseline, but require tuning of \mlr{} to prevent convergence on the guidance objective. In contrast, the learned $\beta_1$ (Figure \ref{fig:beta1_sweep}) runs converge to a similar schedule given a sufficiently high \mlr{}, decaying from high to low momentum over the course of training, regardless of the initialization value. All runs with guided $\beta_1$ outperform the baseline. 
To evaluate how well these gains generalize, we guide $\alpha$ and $\beta_1$ alone and together for 6 language pairs, setting \mlr{} to $3\text{e-}5$ for all runs (Table \ref{tab:combo_bleu}).







\begin{wraptable}{r}{7cm}
  \caption{BLEU scores of baseline vs guided runs across language pairs. \mlr{} is set to $3\text{e-}5$.}
  \footnotesize
  \centering
  \begin{tabular}{c|cccccc}
    \toprule
    & \multicolumn{6}{l}{de-en \hspace{0.17cm} en-de \hspace{0.17cm}  fi-en \hspace{0.17cm}  en-fi \hspace{0.17cm}  lt-en \hspace{0.17cm}  en-lt} \\
    \midrule
    base & 38.6 & 37.4 & 27.2 & 18.4 & 27.3 & 11.3 \\
    \midrule
    $\alpha$ & 39.6 & \textbf{39.4} & 27.6 & \textbf{19.7} & 27.7 & 11.7 \\
    $\beta_1$ & 39.8 & \textbf{39.4} & \textbf{28.4} & 19.4 & \textbf{28.0} & 12.1 \\
    $\alpha$\plus$\beta_1$  & \textbf{39.9} & 38.8 & 27.5 & 19.6 & 27.8 & \textbf{12.2} \\

    \bottomrule
  \end{tabular}
  \vspace{2mm} 
  \label{tab:combo_bleu}
  \caption{BLEU scores of $100$ BO-tuned runs vs un-tuned for baseline and guided runs, on \textit{de-en}. Time is summed runtime in hours.}
  \footnotesize
  \centering
  \begin{tabular}{c|c|c|ccc}
    \toprule
    & time & \# runs & $\alpha$ & $\beta_1$ & $\alpha$\plus$\beta_1$\\
    \midrule
    base & 7.1 & 1 & 38.6 & 38.6 & 38.6 \\
    + BO & 708 &100 & 39.4 & 39.4 & 39.5 \\
    \midrule
    guided & 43.2 & 4 & 39.6 & 39.8 & \textbf{39.9} \\
    + BO & 1.1k &100 & 39.7 & 39.8 & \textbf{39.9} \\
    \bottomrule
  \end{tabular}
  \label{tab:tuned}
\end{wraptable}







In order to evaluate the practical applicability of guiding these hyperparameters, we compare the guided runs to a typical hyperparameter optimization scheme, against which we can evaluate both performance and efficiency. We tune both the baseline runs (via the hyper-parameters directly) and the guided runs (via \mlr{}) with Bayesian optimization (BO)\footnote{The specific algorithm we use is Gaussian Process Bandits \citep{frazier2018tutorial,Vizier}.} for 100 trials. In Table \ref{tab:tuned}, we find that across guided-parameter settings, the non-BO-optimized guided run outperforms the best BO-tuned baseline model, with some slight gains for the guided $\alpha$ run with further BO-tuning\footnote{We count the 4 different values of \mlr{} we tried in Figure \ref{fig:lr_sweep} as tuning runs for the non-BO-tuned guided setup.}. Note the guided $\beta_1$ runs do not require \mlr{} tuning to reach best performance. 

For all setups, the learned hyper-parameters achieve better performance than Bayesian optimization in fewer training runs and less time. Though the `short-horizon bias' requires tuning \mlr{} while learning $\alpha$, doing so still yields performance and efficiency gains over BO-tuning. For $\beta_1$ alone, there seems to be no equivalent bias, as any sufficiently high \mlr{} converges to roughly the same useful schedule. The BO-tuned optimal static $\beta_1$ value (0.73) approximates the average $\beta_1$ of the converged runs in Figure \ref{fig:beta1_sweep}, suggesting that the remaining 0.4 BLEU points are only attainable with a $\beta_1$ value that changes over the course of training. Learning both hyper-parameters together does not change their evolution but yields a small additional boost.

\subsection{T5 pretraining}
\begin{wrapfigure}{r}{0.5\textwidth}
  \begin{center}
    \includegraphics[width=0.4\textwidth]{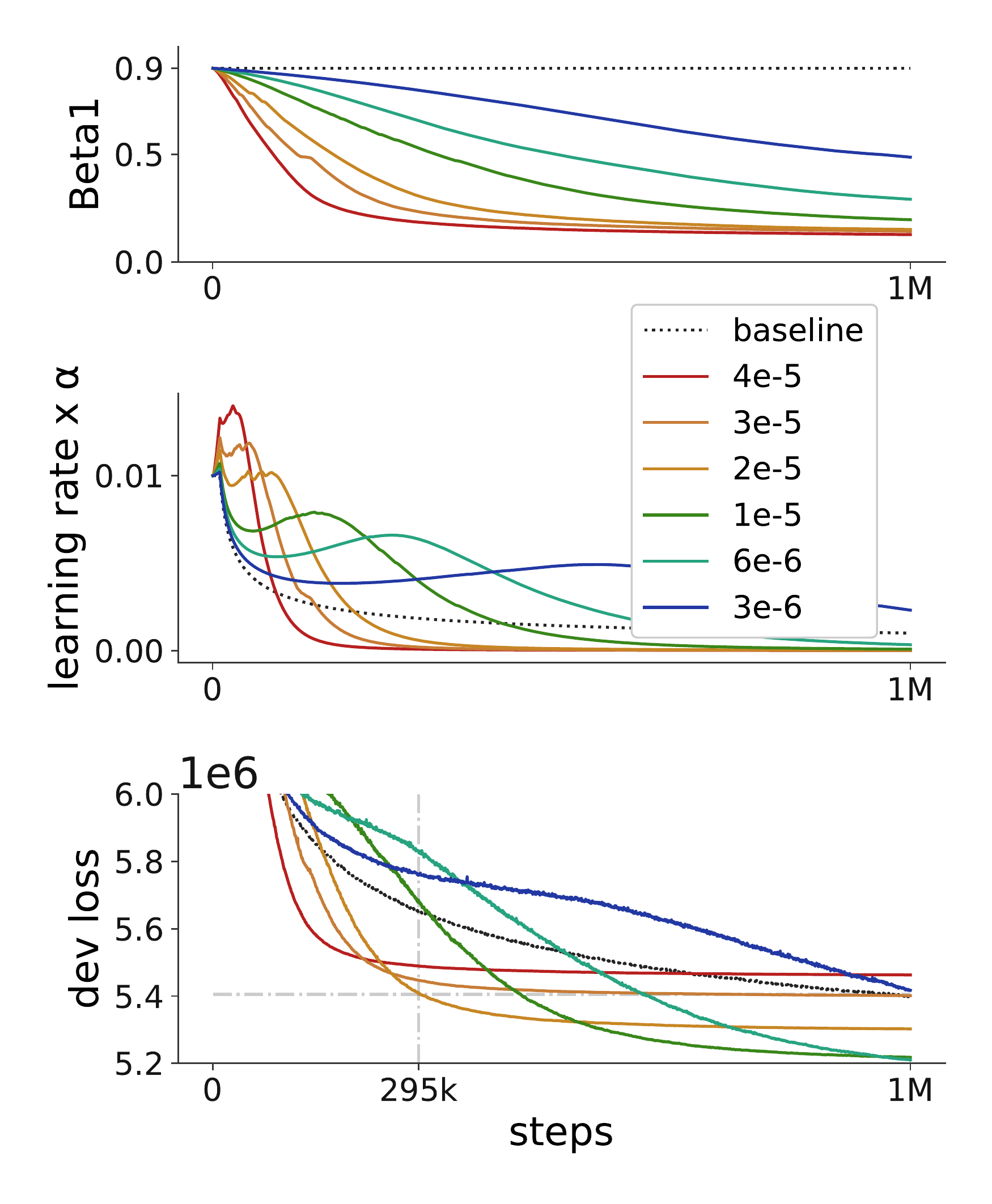}
    \end{center}
    \caption{Learning $\alpha$\plus$\beta_1$ for T5, varying \mlr{}.
    }
	\label{fig:t5}
\end{wrapfigure}
We run similar experiments for T5 models, learning $\alpha$, $\beta_1$, and both. For $\alpha$, we see the learning rate scalar decrease prematurely similarly to the NMT setting, demonstrating again the `short-horizon bias' (Appendix, Figure
2), but no guided run outperforms the baseline, even with low \mlr{} values\footnote{See Appendix \ref{appendix:t5} 
for full details on isolated $\alpha$ and $\beta_1$ experiments.}. For $\beta_1$ alone, we replicate a similar converged schedule as in the NMT setting, but see only minor changes in development set loss across all models, including those varying $\beta_1$ without learning during training (Appendix, Figure
2). This suggests that tuning $\beta_1$ in general is less useful in this setting. Interestingly, when we tune both hyper-parameters together, the evolution of the $\alpha$ parameter changes character (Figure \ref{fig:t5}), and we find a $3$X improvement in speed of convergence relative to baseline and increases in final performance for multiple different settings of \mlr{}. 


We finetune the baseline and \mlr{}=$1\text{e-}5$ models on each of the downstream NLU tasks drawn from the GLUE \citep{wang2018glue} and superGLUE \citep{wang2019superglue} benchmarks, as well as SQuAD \citep{rajpurkar2016squad}, using the same finetuning settings as the original T5 paper\footnote{For details on the setup of finetuning, 
see Appendix \ref{appendix:t5_down}. For
full results, including on SQuAD, see  Appendix \ref{appendix:t5_down_results}.} 
(Table \ref{tab:t5_down}). We find improvements across 15 of 18 downstream NLU tasks, with average improvements of 0.4 points on GLUE and 1.4 points on superGLUE. 


\begin{table*}[b]
    \begin{tabularx}{1\textwidth}{p{0.9cm}|*{9}{X}|c}


    \toprule
    \multirow{2}{*}{GLUE} & CoLA & SST & MRPC & STS & QQP & MNLI & QNLI & RTE & WNLI & \multirow{2}{*}{avg} \\
    & $\phi{}$\textit{corr} & \textit{acc} & \met{} & \met{} & \met{} & \met{} & \textit{acc} & \textit{acc} & \textit{acc} & \\
    \cmidrule(lr){2-11}
    base & \textbf{47.9} & 92.3 & 88.5 & 84.1 & 87.4 & 84.1 & 90.2 & 67.9 & 57.7  & 77.8 \\
    $\alpha$\plus$\beta_1$ & 44.5 & \textbf{92.6} & \textbf{90.0} & \textbf{85.1} & \textbf{87.7} & \textbf{84.9} & \textbf{90.5} & \textbf{69.0} & \textbf{59.2}  & \textbf{78.2} \\
  \end{tabularx}
    \begin{tabularx}{1\textwidth}{p{0.9cm}|*{8}{X}|c}
    \midrule
    \multirow{2}{*}{sGLUE} & BoolQ & CB & COPA & MultiRC & ReCoRD & RTE & WiC & WSC & \multirow{2}{*}{avg}  \\
    
    & \textit{acc} & \met{} & \textit{acc} & \met{} & \met{} & \textit{acc} & \textit{acc} & \textit{acc} & \\
    \cmidrule(lr){2-10}
    base & 73.6 & \textbf{98.4} & 58.0 & 43.7 & 63.8 & 64.6 & 67.2 & 67.3 & 67.1 \\
    $\alpha$\plus$\beta_1$ & 73.6 & 95.3 & \textbf{61.0} & \textbf{46.0} & \textbf{64.0} & \textbf{66.4} & \textbf{67.6} & \textbf{74.0}& \textbf{68.5} \\
    \bottomrule
  \end{tabularx}
  \caption{Fine-tuning baseline and $\alpha$\plus$\beta_1$ models on the GLUE and superGLUE (sGLUE) NLU tasks. \protect\met{} denotes the mean of the two metrics typically reported for that task, and \protect\textit{avg} takes the average across tasks. Max value of 5 runs shown, see Appendix \protect\ref{appendix:t5_down_results} 
  for average results.}

  \label{tab:t5_down}
\end{table*}

\section{Discussion and Limitations}\label{sec:conclusion}


Our results shed light into multiple facets of Hyper-parameter optimization (HPO). For Neural Machine Translation, we show that although learning the learning rate scalar decays the learning rate prematurely when allowed to converge to the guidance objective (exhibiting the `short-horizon bias' \citep{wu2018}), tuning the meta-learning-rate produces better results with less tuning than Bayesian optimized static tuning. In learning momentum, we demonstrate the absence of short-horizon bias; for momentum, and potentially other hyper-parameters, greedy gradient-based HPO can learn over a single run a schedule which out-performs optimal static tuning.
For hyper-parameters such as momentum whose optimal values change over training, methods which allow for dynamic hyper-parameters will always have an edge over static tuning methods. 

In our T5 experiments, we show that the `recipe' which yielded good results in NMT produced, with minimal tuning, a pretrained model which outperforms the baseline after finetuning on downstream NLU tasks. We discovered that learning hyper-parameters in conjunction can alter their evolution over training. When learned alongside momentum, the initial growth of the learning-rate scalar followed by gradual decay is a result that is not explicable by the short-horizon bias, which would predict monotonic and premature decay to zero. This raises the possibility that learning certain hyper-parameters dynamically may be constrained by the static values of non-learned hyper-parameters, and that learning multiple hyper-parameters together may be necessary in some settings to make learning any of them useful. Characterizing the phenomenon of interaction between hyper-parameters  
is a direction for future work.

Our experiments are limited to two global hyper-parameters which are typically tuned. Future work should explore a wider set of hyper-parameters and at a varying granularity (e.g. a distinct hyper-parameter value per parameter \citep{lorraine2020}). We show that learning hyper-parameters together can alter their dynamics but leave to future work the characterization of the mechanism and mapping of interactions between learned hyper-parameters. We have shown that greedily learning the learning rate scalar can produce behavior unexplained by the short-horizon bias, but have left to future work the characterization of this phenomenon. The method we explore is limited to differentiable hyper-parameters, and is greedy, so may be improved upon by more complex methods which can take into account either non-differentiable hyperparameters \citep{mackay2019} and/or longer horizons \citep{micaelli2021}.
 

\section{Broader Impact}\label{sec:impact}
Since \cite{wu2018} described short-horizon bias for greedy methods, work in the gradient-based HPO community has progressed towards more complex methods which seek to address short-horizon bias with longer horizons \citep{micaelli2021} or by other means \citep{donini2019}. Our result showing the absence of bias for learning momentum, and easy performance gains for NMT when doing so, should encourage further evaluation of the behavior of diverse learnable hyper-parameters under greedy meta-optimization. Additionally, we have shown that intuitions about the short-horizon bias do not fully explain the behavior of the learning-rate scalar, which increases at the start of training when learned alongside momentum. These observations, taken together, should encourage further exploration of greedy gradient-based methods.
We do not anticipate this work having potential negative societal impacts beyond those posed by automated methods in machine learning in general.
Rather we hope that it may contribute towards the realization of efficient and general gradient-based HPO, which will help improve the efficiency of training models, reduce energy consumption, and democratize access to machine learning. We hope that our encouraging results and release of the code we used to produce them\footnote{\url{https://www.github.com/google-research/google-research/tree/master/gradient\_based\_tuning}} will facilitate future work within the research community and give practitioners the tools to apply gradient-based HPO in diverse settings.

\section{Reproducibility Checklist}

\begin{enumerate}
\item For all authors\dots
  \begin{enumerate}
  \item Do the main claims made in the abstract and introduction accurately
    reflect the paper's contributions and scope?
    \answerYes{See Sections \ref{sec:experiments} and \ref{sec:conclusion}}.,
  \item Did you describe the limitations of your work?
    \answerYes{See latter portion of Section \ref{sec:conclusion}.}
  \item Did you discuss any potential negative societal impacts of your work?
    \answerNA{We anticipate no specific potential negative impacts beyond those of improving automated machine learning methods in general. We state this in Section \ref{sec:impact}.}
  \item Have you read the ethics author's and review guidelines and ensured that your paper
    conforms to them? \url{https://automl.cc/ethics-accessibility/}
    \answerYes{We do not violate the guidelines.}
  \end{enumerate}
\item If you are including theoretical results\dots
  \begin{enumerate}
  \item Did you state the full set of assumptions of all theoretical results?
    \answerNA{We present no theoretical results.}
  \item Did you include complete proofs of all theoretical results?
    \answerNA{We present no theoretical results.}
  \end{enumerate}
\item If you ran experiments\dots
  \begin{enumerate}
  \item Did you include the code, data, and instructions needed to reproduce the
    main experimental results, including all requirements (e.g.,
    \texttt{requirements.txt} with explicit version), an instructive
    \texttt{README} with installation, and execution commands (either in the
    supplemental material or as a \textsc{url})?
    \answerNA{We will release the code prior to the publication of the work. While this is clearly not the same as releasing it now (at submission time), we intend to do so as open-sourcing the code is a main aspect of the intended impact of the work.}
  \item Did you include the raw results of running the given instructions on the
    given code and data?
    \answerNA{See above.}
  \item Did you include scripts and commands that can be used to generate the
    figures and tables in your paper based on the raw results of the code, data,
    and instructions given?
    \answerNA{Close analogues of the figures in this paper will be automatically generated by the training code.}
  \item Did you ensure sufficient code quality such that your code can be safely
    executed and the code is properly documented?
    \answerYes{The code, which will be released prior to publication, will be well documented.}
  \item Did you specify all the training details (e.g., data splits,
    pre-processing, search spaces, fixed hyperparameter settings, and how they
    were chosen)?
    %
    \answerYes{See Appendix \ref{appendix:setup}.}
  \item Did you ensure that you compared different methods (including your own)
    exactly on the same benchmarks, including the same datasets, search space,
    code for training and hyperparameters for that code?
    \answerYes{We took care to ensure our experiments comparing methods were fair, including in these mentioned categories.}
  \item Did you run ablation studies to assess the impact of different
    components of your approach?
    \answerYes{We vary \mlr{}, combination of hyper-parameters, and in comparing NMT to T5 pretraining, we vary optimizer, model, and task.}
  \item Did you use the same evaluation protocol for the methods being compared?
    %
    \answerYes{See section \ref{sec:experiments} and Appendix \ref{appendix:setup}.}
  \item Did you compare performance over time?
    \answerYes{See Figures in Section \ref{sec:experiments}.}
  \item Did you perform multiple runs of your experiments and report random seeds?
    \answerNo{We did perform multiple runs of the experiments but do not report random seeds.}
  \item Did you report error bars (e.g., with respect to the random seed after
    running experiments multiple times)?
    \answerNo{We do not report error bars, we report the max and average metric values over repeated runs.}
  \item Did you use tabular or surrogate benchmarks for in-depth evaluations?
    \answerNA{We do not employ NAS approaches.}
  \item Did you include the total amount of compute and the type of resources
    used (e.g., type of \textsc{gpu}s, internal cluster, or cloud provider)?
    %
    \answerYes{See Appendix \ref{appendix:hardware}.}
  \item Did you report how you tuned hyperparameters, and what time and
    resources this required (if they were not automatically tuned by your AutoML
    method, e.g. in a \textsc{nas} approach; and also hyperparameters of your
    own method)?
    \answerYes{See Section \ref{sec:experiments}.}
  \end{enumerate}
\item If you are using existing assets (e.g., code, data, models) or
  curating/releasing new assets\dots
  \begin{enumerate}
  \item If your work uses existing assets, did you cite the creators?
    \answerYes{See Section \ref{sec:experiments}.}
  \item Did you mention the license of the assets?
    %
    \answerYes{See Appendix \ref{appendix:license}.}
  \item Did you include any new assets either in the supplemental material or as
    a \textsc{url}?
    \answerNo{We will include a link to the code at publication time.}
  \item Did you discuss whether and how consent was obtained from people whose
    data you're using/curating?
    \answerNA{Our experiments were performed on publicly available datasets.}
  \item Did you discuss whether the data you are using/curating contains
    personally identifiable information or offensive content?
    \answerNo{None of our datasets contains personally identifiable information or offensive content.}
  \end{enumerate}
\item If you used crowdsourcing or conducted research with human subjects\dots
  \begin{enumerate}
  \item Did you include the full text of instructions given to participants and
    screenshots, if applicable?
    \answerNA{}
  \item Did you describe any potential participant risks, with links to
    Institutional Review Board (\textsc{irb}) approvals, if applicable?
    \answerNA{}
  \item Did you include the estimated hourly wage paid to participants and the
    total amount spent on participant compensation?
    \answerNA{}
  \end{enumerate}
\end{enumerate}

\begin{acknowledgements}
The authors would like to thank Andrew Chou, Felix Stahlberg, Ji Ma,
and the anonymous reviewers, for their helpful comments.
\end{acknowledgements}


\bibliography{example_paper}
\bibliographystyle{icml2021}
\newpage
\appendix

\section{Related Work}\label{appendix:related_work}
The field of hyperparameter optimization (HPO) is well summarized in \citet{Feurer2019}. Here we review related work in gradient-based HPO, of which our method is one approach. Online gradient-based HPO was proposed by \citet{almeida1998parameter}. \citet{Bengio2000} formulated hyperparameter search in terms of optimization. \citet{domke2012}  described a strategy to compute the gradient of loss with respect to hyperparameters in a CRF model. The use of validation-loss gradients to update continuous hyperparameters by backpropagating through the entire training procedure was demonstrated by \citet{maclaurin15}. To reduce time-complexity of tracing back through the entire training procedure, subsequent work explored approaches where the parameter and hyperparameter updates are performed in an alternating fashion \citep{luketina2016scalable,franceschi_forwardreverse_2017,franceschi_bilevel_2018,baydin2017online,majumdar2019}. \citet{luketina2016scalable} proposed greedy per-step validation loss gradient updates, applied to regularization hyperparameters that are trained alongside the elementary parameters of the model. \citet{baydin2017online} described an application of the greedy approach to optimize learning rates using the training set loss. \citet{wu2018} highlighted the short-horizon biases arising from the greedy strategy. \citet{Fu2016,donini2019,micaelli2021} presented approaches that overcome some of the limitations of the greedy strategy while being more efficient than the full trajectory approach of \citet{maclaurin15}.  The above methods considered either forward- or reverse-mode differentiation to compute the hyper-gradients. Alternative approaches, using the Implicit Function Theorem to approximate the gradients, were explored in \citet{pedregosa2016, lorraine2020, clarke2021,grazzi2021}. \citet{shaban2019truncated} proposed an approach using truncated backpropagation to approximate the hypergradient. \citet{mackay2019} presented a method for learning a hyperparameter schedule that works for non-differentiable hyperparameters. Some works focus on using gradients to learn data weighting or augmentation schemes, such as \cite{hu2019learning}. \cite{raghu2020teaching} leverages gradient methods to learn various 
`commentaries' that are example-level parameters that can improve performance via example weighting and data manipulation and also provide insights into model training. MAML \citep{finn2017model} and subsequent works \citep{antoniou2018train,bansal2020self} employ a bi-level, gradient based training procedure using a distribution over tasks that improves the generalization performance and can be utilized to learn hyperparameters. \cite{raghu2021meta} apply a gradient-based method to meta-learn hyperparameters for multi-task pretraining on protein-protein interaction networks.







\section{Algorithm}\label{appendix:algorithm}
\begin{algorithm}
\caption{Guided Learning}
\begin{algorithmic}
\Require $\theta_0$: initial parameter vector 
\Require $\lambda_0$: initial hyperparameter vector
\Require $\hat{\lambda}$: hyperparameter vector of meta-optimizer
\State $t \leftarrow 0$\algorithmiccomment{Initialization}
\While {$\theta_t$ not converged}
\State $(X_t, H_t) \leftarrow \text{GetNewMiniBatch}()$ 
\algorithmiccomment{New training/guidance mini-batch}
\State $g^{X}_t \leftarrow \nabla_{\theta_{t}}\text{ComputeLoss}(X_t, \theta_{t})$
\algorithmiccomment {Gradient of train loss wrt $\theta_{t}$}
\State $\theta_{t+1} \leftarrow \text{Optimizer}(g^{X}_{t}, \theta_{t}, \lambda_{t})$ \algorithmiccomment{Parameter update}
\State $\hat{g}^{H}_t \leftarrow \nabla_{\lambda_{t}} \text{ComputeLoss}(H_t, \theta_{t+1})$
\algorithmiccomment {Gradient of guidance loss wrt $\lambda_{t}$ }
\State $\lambda_{t+1} \leftarrow \text{MetaOptimizer}(\hat{g}^{H}_{t}, \lambda_{t}, \hat{\lambda})$
\algorithmiccomment{Hyperparameter update}
\State $t \leftarrow t+1$
\EndWhile
\end{algorithmic}
\label{algo:guided_learning}
\end{algorithm}

\section{Setup}\label{appendix:setup}

\subsection{NMT Experiments}

\subsubsection{Model}
For all experiments, we use the JAX framework \citep{jax2018github}, building off of models from the flax library \citep{flax2020github}. We use Transformer models \citep{vaswani2017attention} and the LAMB optimizer \citep{lamb}, with a 32k sentence-piece vocabulary \citep{kudo-richardson-2018-sentencepiece} for each language pair. Our Transformers have 8 heads and 6 layers with a total of 121M parameters, and for the LAMB optimizer we use the default values of $\beta_1$, $\beta_2$, and $\epsilon$ as $0.9$, $0.999$, and $1\text{e-}6$ respectively.

\subsubsection{Data}
We train on $6$ different language pairs, with training, development, and test sets drawn from the WMT19 machine translation task \citep{barrault-etal-2019-findings}. We tokenize the language pairs into joint 32K subword vocabularies with SentencePiece models \citep{kudo-richardson-2018-sentencepiece}. After filtering the datasets slightly by language ID and with length-based heuristics, we remove a single batch of the remaining data to set aside as a guidance set for each language pair. This is based on our preliminary experiments where we found no change in performance between holding out 1\% of the training data for guidance (iterated through repeatedly over training) or holding out a single batch (applied at every step), so throughout this work we hold out only a single batch for the guidance set\footnote{We likely see no difference because at most we learn two hyperparameters. With higher-dimensional learned hyperparameterizations, overfitting on the guidance set may become a concern that can be addressed by iterating through a larger guidance dataset.}. 
The resulting dataset sizes are shown in Table \ref{tab:data}.

\begin{table}[h]
  \footnotesize
  \caption{Comparing dataset sentence count across language pairs. The acronyms de, en, fi and lt refer to German, English, Finnish and Lithuanian, respectively.}
  \centering
  \begin{tabular}{c|cccccc}
    \toprule
    & de-en & en-de & fi-en & en-fi & lt-en & en-lt \\
    \midrule
    train & 32M & 32M & 5.5M & 5.5M & 1.9M & 1.9M \\
    guide & 2165 & 2227 & 2363 & 2337 & 2339 & 2305 \\
    \bottomrule
  \end{tabular}
  \label{tab:data}
\end{table}

\subsection{Training}
We train with dropout and attention dropout both set to $0.1$, and without label smoothing or weight decay regularization. The default learning rate is set to $0.4$, which follows a square-root decay schedule following a a linear warmup of $4000$ steps. We use a training batch size of  ${\sim}2,300$ examples on average. In the experiments where we compare learned hyperparameters to Bayesian HPO \citep{Snoek12}, the objective for the BO is to minimize the loss on the development set, and we select the best of 100 trials for each BO run.
 
\subsubsection{Evaluation}
We decode with beam search decoding with a beam size of 4, and report BLEU \citep{papineni-etal-2002-bleu} scores calculated using the sacreBLEU tool \citep{post-2018-sacrebleu}.

\subsection{T5 Experiments}

\subsubsection{Model}
For the T5 experiments, we pretrain \textit{T5} models \citep{raffel2019exploring} using the \textit{Adafactor} optimizer \citep{shazeer2018adafactor}. We train a T5 model in the \textit{small} configuration, with 8 layers and 6 attention heads per layer and a total of 60M parameters. For Adafactor we use a default learning rate of $1\text{e-}3$ and a \textit{decay\_rate} of 0.8. 

\subsubsection{Data and Training}
For pretraining, we train for 1M steps on the C4 dataset \citep{raffel2019exploring}, using a 32k sentence-piece vocabulary, the same as in the original T5 paper. We use a batch size of 256 packed examples and a maximum input length of 512 sentence-pieces, with dropout set to 0.0. We use a learning rate of 0.01 with 10000 steps of constant value followed by reciprocal square root decay. The unsupervised objective is the same masked language modeling objective that was proposed in the original T5 paper. 15\% of tokens are masked in the input sequence, replacing each masked span with a sentinel token. The model is then trained to predict the missing text for each sentinel token. 

\subsubsection{Evaluation}
In pretraining, we report the loss on the C4 development set. For finetuning, we evaluate the appropriate metrics for each of the GLUE and superGLUE tasks. To arrive at the final average for each set of tasks, we follow the T5 paper in averaging the metrics within each tasks (to get the \met{} values shown in Table \ref{tab:t5_down}) and then simply averaging those scores across the tasks of the super-task.

\subsection{Finetuning on Downstream NLU Tasks}\label{appendix:t5_down}

We finetune on downstream NLU tasks from the GLUE and superGLUE meta-tasks. We initialize from the 1M step pretraining checkpoints and train for an additional 250,000 steps with a batch size of 8, mirroring the T5 paper finetuning scheme \citep{raffel2019exploring}.

\subsection{Meta-Optimization}
In our hyperparameter learning experiments, we meta-optimize with Adam and its default hyperparameters. $\beta_1$, $\beta_2$, and $\epsilon$ are set to $0.9$, $0.999$, and $1\text{e-}8$ respectively. For both NMT and T5 experiments, we use a guidance batch size mirroring the size of the training batch in each setting. 

While model parameters may be allowed to take positive or negative values, the hyperparameters we study must be bound to a range of appropriate values; the learning rate must be positive and momentum must be between 0 and 1. To achieve this, we pass the learned hyperparameters through an activation function; exponential for learning rate and sigmoid for momentum (Table \ref{tab:activation_functions}). Unlike other hyperparameters, the learning rate is frequently set on a pre-determined schedule. In order to not override the pre-existing schedule, we learn a scalar $\alpha$ on the schedule which is initialized at 1.

\begin{table}[h]
  \footnotesize
  \caption{Learned hyperparameters and their activation functions.}
  \centering
  \begin{tabular}{c|c|c|c}
    \toprule
    hparam & activation fn & domain & init \\
    \midrule
    $\alpha$ & $e^x$ & (0, $\infty$) & 1 \\
    $\beta_1$ & $(1+e^{-x})^{-1}$ & (0, 1) & 0.9 \\
    \bottomrule
  \end{tabular}
  \label{tab:activation_functions}
\end{table}

\section{Hardware}\label{appendix:hardware}
For all experiments, we use TPUv3 with 16 cores. NMT training runs took $\sim$7 hours to train. T5 training took $\sim$48 hours for pretraining and $\sim$3-6 hours for finetuning depending on the task. In both setups, training runs that guided hyper-parameters took approximately 1.5X as long in terms of wall-clock time than baseline runs. The memory requirements of the guided and unguided runs were similar.

\pagebreak
\section{Tuning $\beta_1$ via Bayesian optimization}\label{appendix:tuning_beta1}

\setcounter{figure}{0}
\begin{figure}[!ht]
	\centering%
    \includegraphics[width=0.5\linewidth]{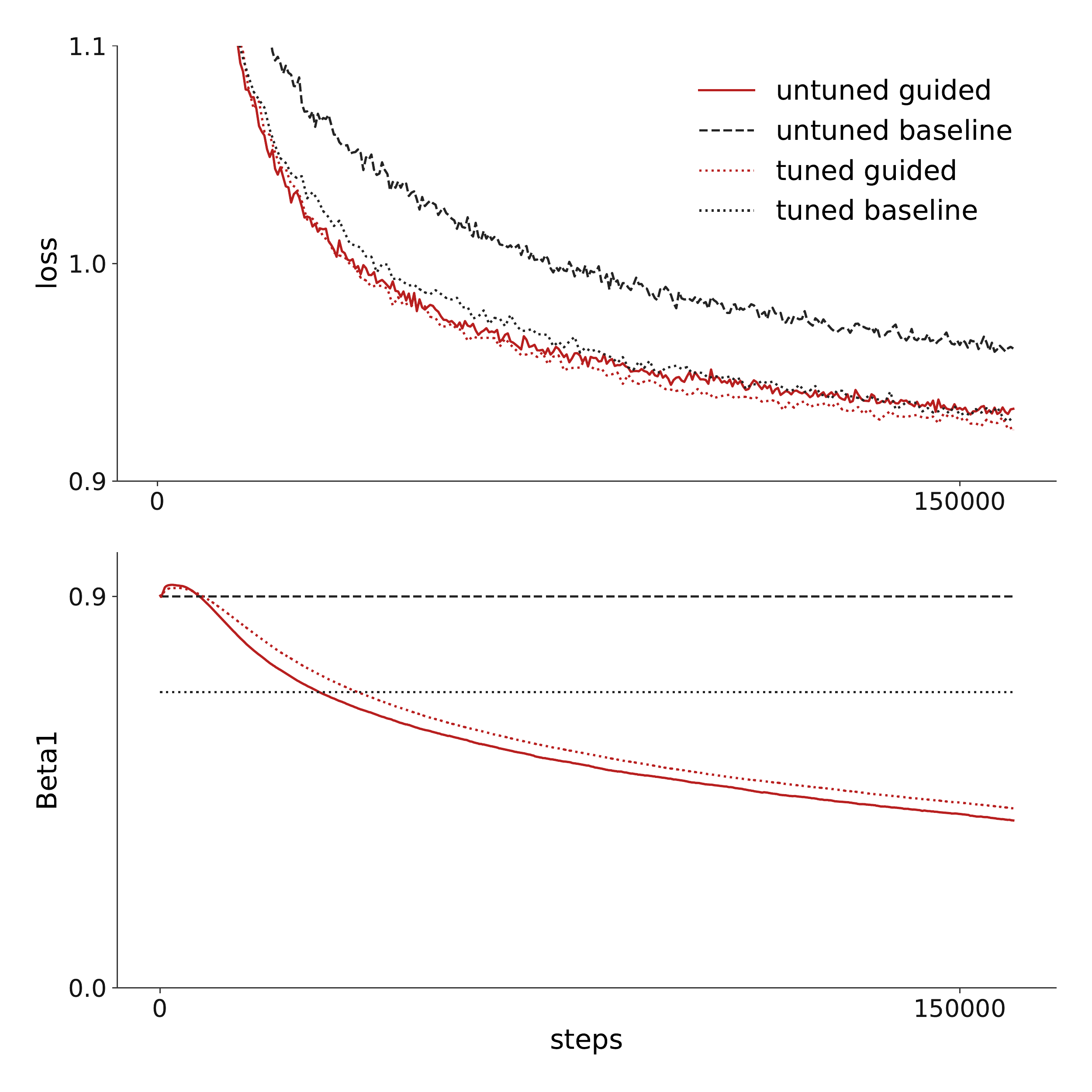}
        \caption{Learning $\beta_1$ for NMT models, comparison of untuned learned and baseline runs (solid lines) to BO-tuned learned and baseline runs (dotted lines). These runs correspond to those reported in the $\beta_1$ column in Table \ref{tab:tuned}. 
        }
	\label{fig:draft}
\end{figure}

\pagebreak

\section{T5 experiments}\label{appendix:t5}

\begin{figure}[!ht]
	\centering%
    \includegraphics[width=0.5\linewidth]{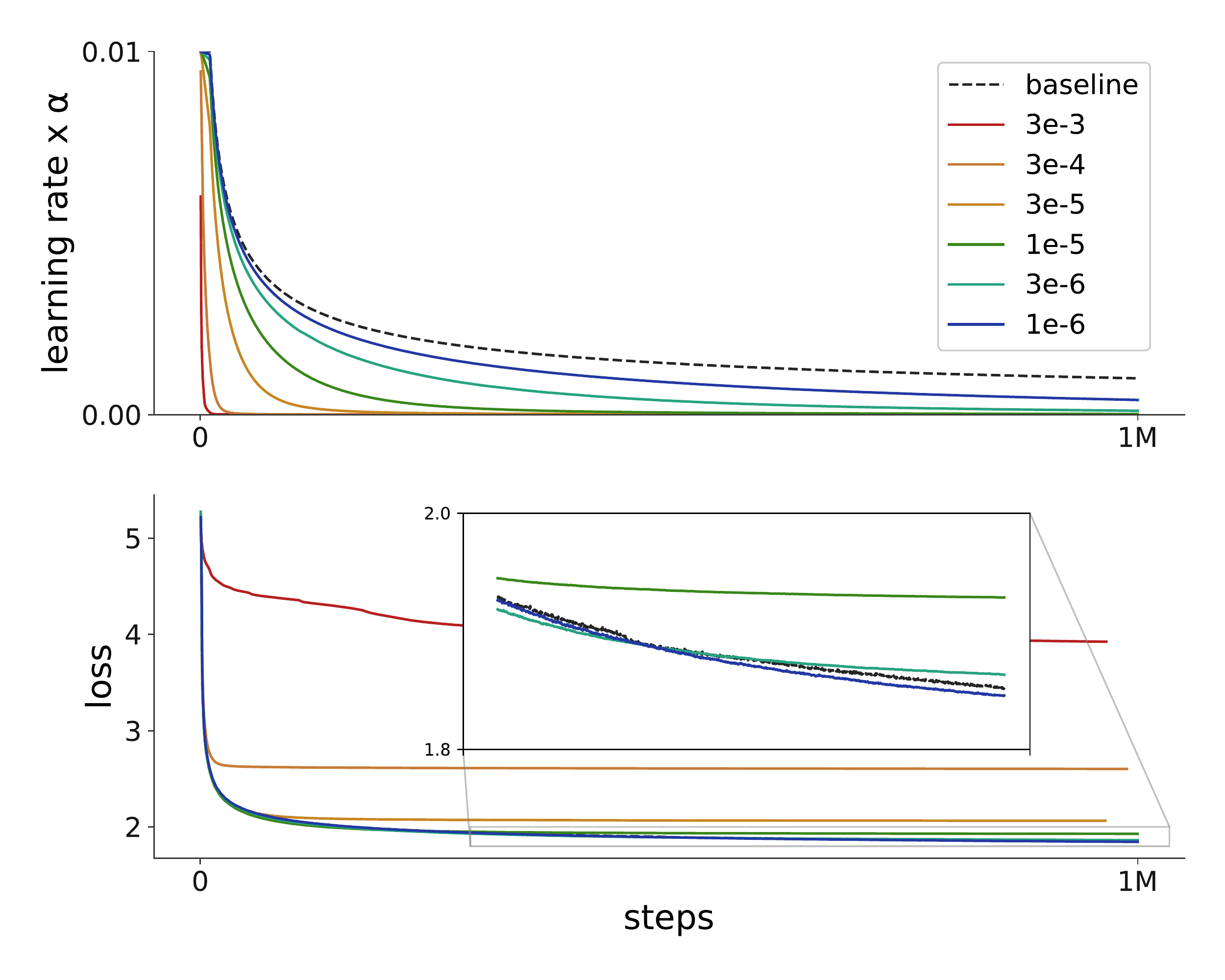}
        \caption{Learning $\alpha$ alone for T5 pretraining, comparison of a sweep over color-coded meta learning rates to baseline (black). The lowest meta-learning rate setting ($1\text{e-}6$, in blue) does outperform the baseline, but only very slightly. The short-horizon bias is evident; note that all learning rate scalars only decrease relative to the baseline learning rate schedule.
        }
	\label{fig:t5alpha}
\end{figure}

\begin{figure}[!hb]
	\centering%
    \includegraphics[width=0.5\linewidth]{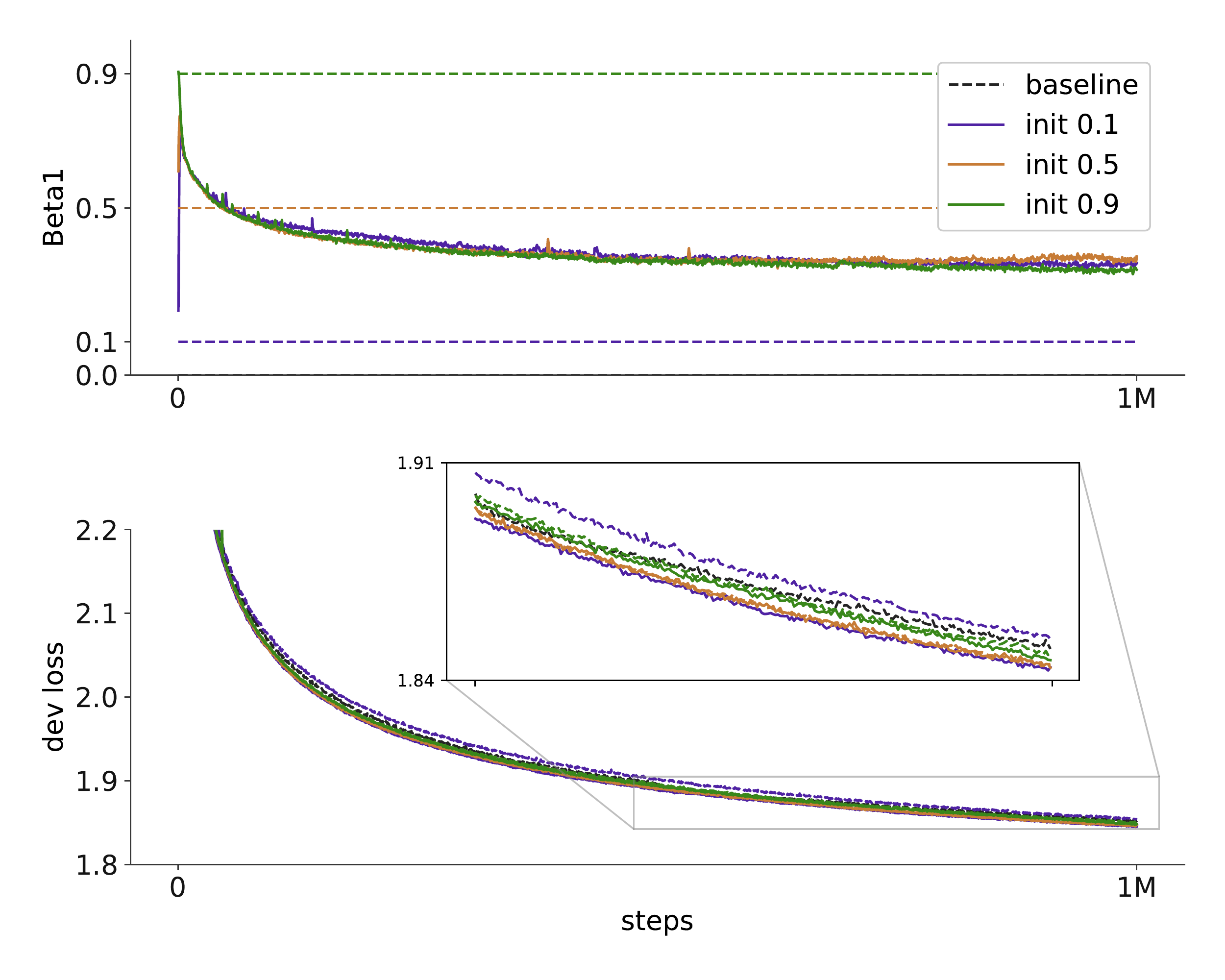}
        \caption{Learning $\beta_1$ alone for T5 pretraining, comparison of learned (meta learning rate $3\text{e-}3$) runs vs baseline for initializations [$0.1$, $0.5$, $0.9$], and default baseline $0.0$. Note that learned $\beta_1$ values converge to the same gradually decaying schedule, similar to that of the NMT models in Figure \ref{fig:beta1_sweep}. The runs on that schedule do very slightly out-perform the non-learned hyperparameter runs. However, unlike in the NMT case, none of the changes in $\beta_1$, dynamic or static, have a significant impact upon the accuracy of the model at any point in training. This suggests that this setup is simply insensitive to the value of $\beta_1$.
        }
	\label{fig:t5beta}
\end{figure}

\pagebreak

\subsection{Downstream NLU Task Full Results}\label{appendix:t5_down_results}

\begin{table}[h!]
\centering
  \begin{tabular}{p{0.9cm}|*{8}{c}}
    
    \toprule
    & GLUE & CoLA & SST & MRPC & MRPC & STS & STS & QQP \\ 
    & avg & $\phi{}$\textit{corr} & \textit{acc} & \textit{F1} & \textit{acc} & \textit{Pearson} & \textit{Spearman} & \textit{F1} \\
    \cmidrule(lr){2-9}
    base & 77.8 & 47.9 & 92.3 & 90.3 & 86.7 & 84.0 & 84.3 & 85.6 \\ 
    $\alpha$\plus$\beta_1$ & 78.2 & 44.5 & 92.5 & 91.7 & 88.2 & 84.9 & 85.3 & 86.0 \\
    \midrule
     & QQP & MNLI-m & MNLI-mm & QNLI & RTE & WNLI & SQuAD & SQuAD \\
    & \textit{acc} & \textit{acc} & \textit{acc} & \textit{acc} & \textit{acc} & \textit{acc} & \textit{EM} & \textit{F1}  \\
    \cmidrule(lr){2-7}
    \cmidrule(lr){8-9}
    base & 89.2 & 83.8 & 84.5 & 90.2 & 67.9 & 57.7 & 88.2 & 80.1  \\
    $\alpha$\plus$\beta_1$ & 89.5 & 84.6 & 85.2 & 90.5 & 69.0 & 59.2 & 88.3 & 80.6 \\
    \toprule
    & sGLUE & BoolQ & CB & CB & COPA & MultiRC & MultiRC & ReCoRD \\
    & avg & \textit{acc} & \textit{F1} & \textit{acc} & \textit{acc} & \textit{F1a} & \textit{EM} & \textit{F1} \\
    \cmidrule(lr){2-9}
    base & 67.1 & 73.6 & 98.1 & 98.7 & 58.0 & 18.9 & 68.5 & 63.3 \\
    $\alpha$\plus$\beta_1$ & 68.5 & 73.6 & 94.6 & 96.0 & 61.0 & 22.6 & 69.7 & 63.6 \\

    \midrule
    & ReCoRD & RTE & WiC & WSC & & & & \\
    & \textit{acc} & \textit{acc}  & \textit{acc}  & \textit{acc}  & & & & \\
    \cmidrule(lr){2-9}
    base & 64.3 & 64.6 & 67.1 & 67.3 \\
    $\alpha$\plus$\beta_1$ & 64.5 & 66.4 & 67.6 & 74.0 \\
    \toprule
  \end{tabular}
  \caption{Fine-tuning baseline and learned $\alpha$\plus$\beta_1$ models on SQuAD, plus the 9 GLUE and 8 superGLUE (sGLUE) downstream NLU tasks. All values are the \textit{max} of 5 separate runs.}
  \label{tab:t5_down_results_a}
\end{table}

\begin{table}[h!]
\centering
  \begin{tabular}{p{0.9cm}|*{8}{c}}
    
    \toprule
    & GLUE & CoLA & SST & MRPC & MRPC & STS & STS & QQP \\ 
    & avg & $\phi{}$\textit{corr} & \textit{acc} & \textit{F1} & \textit{acc} & \textit{Pearson} & \textit{Spearman} & \textit{F1} \\
    \cmidrule(lr){2-9}
    base & 79.8 & 47.3 & 92.2 & 89.9 & 86.0 & 83.6 & 84.0 & 87.2 \\ 
    $\alpha$\plus$\beta_1$ & 80.1 & 43.6 & 92.5 & 90.8 & 87.0 & 84.6 & 85.0 & 87.2 \\
    \midrule
     & QQP & MNLI-m & MNLI-mm & QNLI & RTE & WNLI & SQuAD & SQuAD \\
    & \textit{acc} & \textit{acc} & \textit{acc} & \textit{acc} & \textit{acc} & \textit{acc} & \textit{EM} & \textit{F1}  \\
    \cmidrule(lr){2-7}
    \cmidrule(lr){8-9}
    base & 90.5 & 83.8 & 84.4 & 89.1 & 64.8 & 57.5 & 88.1 & 80.0  \\
    $\alpha$\plus$\beta_1$ & 90.5 & 84.5 & 85.1 & 90.3 & 66.0 & 57.2 & 88.2 & 80.5 \\
    \toprule
    & sGLUE & BoolQ & CB & CB & COPA & MultiRC & MultiRC & ReCoRD \\
    & avg & \textit{acc} & \textit{F1} & \textit{acc} & \textit{acc} & \textit{F1a} & \textit{EM} & \textit{F1} \\
    \cmidrule(lr){2-9}
    base & 64.8  & 73.1 & 97.9 & 98.4 & 56.4 & 6.4 & 62.1 & 63.1 \\
    $\alpha$\plus$\beta_1$ & 67.3 & 72.7 & 93.2 & 93.1 & 57.2 & 21.0 & 69.4 & 63.3 \\

    \midrule
    & ReCoRD & RTE & WiC & WSC & & & & \\
    & \textit{acc} & \textit{acc}  & \textit{acc}  & \textit{acc}  & & & & \\
    \cmidrule(lr){2-9}
    base & 64.1 & 63.8 & 66.1 & 63.5 \\
    $\alpha$\plus$\beta_1$ & 64.3 & 65.9 & 66.7 & 73.5 \\
    \toprule

  \end{tabular}
  \caption{Fine-tuning baseline and learned $\alpha$\plus$\beta_1$ models on SQuAD, plus the 9 GLUE and 8 superGLUE (sGLUE) downstream NLU tasks. All values are the \textit{average} of 5 separate runs.}
  \label{tab:t5_down_results_b}
\end{table}

\section{Licensing of Data}\label{appendix:license}
WMT (Workshop on Machine Translation) 2019: 
\textit{http://www.statmt.org/wmt19/translation-task.html} and downloaded from \textit{https://www.tensorflow.org/datasets/catalog/wmt19\_translate}

\begin{quote}
LICENSING OF DATA (from the statmt.org website) states that it can be used for research purposes:

\textit{The data released for the WMT19 news translation task can be freely used for research purposes, we just ask that you cite the WMT19 shared task overview paper, and respect any additional citation requirements on the individual data sets. For other uses of the data, you should consult with original owners of the data sets.}
\end{quote}

The C4 dataset is released by Google, available at 
\textit{https://www.tensorflow.org/datasets/catalog/c4}, licensed by 
\textit{Creative Commons Attribution 4.0 License}



\appendix

\end{document}